\documentclass{article}

\usepackage{PRIMEarxiv}

\usepackage[utf8]{inputenc} 
\usepackage[T1]{fontenc}    
\usepackage{hyperref}       
\usepackage{url}            
\usepackage{booktabs}       
\usepackage{amsfonts}       
\usepackage{amssymb,amsmath,amsthm,array}
\usepackage{bbm}
\usepackage{nicefrac}       
\usepackage{microtype}      
\usepackage{lipsum}
\usepackage{fancyhdr}       
\usepackage{graphicx}       
\usepackage{color}
\usepackage{subfigure}
\usepackage{multirow}
\usepackage{svg}

\title{Probabilistic forecasts of wind power generation in regions with complex topography using deep learning methods: An Arctic case}

\author{
  Odin Foldvik Eikeland \\
  Department of Physics and Technology \\
  UiT-the Arctic University of Norway  \\
  9037 Tromsø, Norway\\
  \texttt{odin.f.eikeland@uit.no} \\
   \And
  Finn Dag Hovem \\
  Ishavskraft Power Company \\
  9024 Tromsø, Norway \\
   \And
  Tom Eirik Olsen \\
  Ishavskraft Power Company \\
  9024 Tromsø, Norway \\
  \And
  Matteo Chiesa\\
  Department of Physics and Technology \\
  UiT-the Arctic University of Norway  \\
  9037 Tromsø, Norway\\
  \texttt{matteo.chiesa@uit.no} \\
   \And
  Filippo Maria Bianchi\thanks{Corresponding author} \\
  Department of Mathematics and Statistics \\
  UiT-the Arctic University of Norway \\
  9037 Tromsø, Norway\\
  \texttt{filippo.m.bianchi@uit.no} \\
}

\begin{document}
\maketitle
\begin{abstract}
    The energy market relies on forecasting capabilities of both demand and power generation that need to be kept in dynamic balance. Today, when it comes to renewable energy generation, such decisions are increasingly made in a liberalized electricity market environment, where future power generation must be offered through contracts and auction mechanisms, hence based on forecasts. The increased share of highly intermittent power generation from renewable energy sources increases the uncertainty about the expected future power generation. Point forecast does not account for such uncertainties. To account for these uncertainties, it is possible to make probabilistic forecasts.
    
    This work first presents important concepts and approaches concerning probabilistic forecasts with deep learning. Then, deep learning models are used to make probabilistic forecasts of day-ahead power generation from a wind power plant located in Northern Norway. The performance in terms of obtained quality of the prediction intervals is compared for different deep learning models and sets of covariates.   
    
    The findings show that the accuracy of the predictions improves when historical data on measured weather and numerical weather predictions (NWPs) were included as exogenous variables. This allows the model to auto-correct systematic biases in the NWPs using the historical measurement data. Using only NWPs, or only measured weather as exogenous variables, worse prediction performances were obtained.
    
\end{abstract}


\newpage
\section{Introduction}
\label{sec:intro}
Making accurate predictions is of fundamental importance in the energy market where decisions are based on expectations about the future. Today, when it comes to renewable energy generation, such decisions are increasingly made in a liberalized electricity market environment, where future power generation must be offered through contracts and auction mechanisms, hence based on forecasts \cite{liu2020probabilistic,huang2020improved,ruhnau2020economic}. 
Since renewable energy sources (RES) are to eventually participate in market mechanisms under the same rules as conventional fossil-fuel-based generators, mismatches between contracted generation and actual deliveries may induce financial penalties \cite{mazzi2017wind}.
Indeed, the energy production from RES can be predicted with limited accuracy. This, in addition to uncertainties in market prices, yield uncertain market returns. However, in market environments under such high levels of uncertainty, the relevant stakeholders may make better decisions if they are given the best possible estimates about the future. 

Forecasts in their most common form are to predict the next value that is most likely to occur. Forecasts in such form are called a \emph{point} forecast. The main objective of making point forecasts is to train a model to predict a certain point in the future, and hopefully, the actual value will eventually be the same (i.e., a perfect prediction). However, it is not reasonable to expect that a model can perform 100\% accurate predictions. The models will therefore predict the most likely value based on what it has learned through the information that is given.
However, when predicting the most likely value (or point) in the future, one does not consider the uncertainties in this forecast (how sure are we that this will be the next value).
For the stakeholders that work in the energy market with a high degree of uncertainty, it is of interest to measure the uncertainty in a given forecast. 
For instance, the power production from solar photovoltaic (PV) or wind power is highly intermittent and dependent on multiple features with a complex and non-linear nature (weather, power market, human activity) \cite{ahmed2016ensemble,brusaferri2019bayesian,zhang2020improved,liu2020probabilistic}. Making accurate point predictions from these technologies is almost impossible, especially when considering short-term forecasts that range from 12-hour to 36-hour ahead, which is the day-ahead market the energy trading companies must relate to \cite{brusaferri2019bayesian,toubeau2018deep}.
Therefore, it is of fundamental importance to be able to predict the \emph{distribution} of the expected outcome to find a certain interval of possible outcomes. Probabilistic forecasts predict such distributions where the interval of possible outcomes creates a prediction interval (PI). Probabilistic forecasts will allow market participants to consider the uncertainties in a given prediction. 

The increased share of RES technologies with fluctuating generation in the electricity market, and the rapid development within machine learning (ML), have motivated the development of research concerning probabilistic forecasts in energy applications with deep learning. Efforts have been done to construct deep learning models that create probabilistic forecasts of expected generation from technologies such as PV and wind power \cite{huang2020improved,wang2017deterministic,zhang2020improved,wang2017deep,jin2021probabilistic}. Similar approaches have been used to make probabilistic forecasts of expected weather (such as wind speed), traffic, energy load, and spot prices in the electricity market \cite{xiang2020deterministic,salinas2020deepar,zhang2018improved,brusaferri2019bayesian}. The majority have used different deep learning architectures such as Convolutional Neural Networks (CNNs) or Recurrent Neural Networks (RNNs) where different loss functions are optimized to compute predicted distributions. In addition, some works are proposing methodologies where original Neural Networks (NNs) are modified for a specific task to make probabilistic forecasts \cite{afrasiabi2020advanced,toubeau2018deep,wang2017deep}. In \cite{mashlakov2021assessing} the authors compared deep learning models to make probabilistic forecasts on different datasets and showed that wind power generation was the most challenging one to predict with high accuracy due to the highly non-linear feature and randomness in power generation. 
Despite a large number of works on probabilistic forecasts with deep learning techniques in energy applications, there is a lack of research comparing the effect of using exogenous variables to obtain the best prediction performance of wind power generation. 

The electricity generation from wind power systems is directly affected by the amount of wind that hits the turbines, where the amount of power that is generated follows a wind-to-power conversion process, or the \emph{power curve} \cite{zhang2014review}. Therefore, the majority of work uses numerical weather predictions (NWPs) as inputs to forecast the expected wind power generation in the short-term range (6-72 hours ahead) \cite{monteiro2009wind,zhang2020short,higashiyama2018feature,wang2011review}. When predicting wind power generation, a large part of the forecast error will therefore directly come from the NWPs. Predicting short-term weather (especially wind speed) is a complex task due to its non-linear and fluctuating nature \cite{kavasseri2009day,zhao2016one,wang2016analysis}. Consequently, this gives large sources of error for making accurate short-term predictions of wind power generation. The wind power case study considered in this work is in a region with complex terrain where it is large weather variations within small distances, making it increasingly difficult to have accurate NWPs available.
In this work, we address the issue of inaccurate NWPs by constructing different configurations of datasets that could account for such failure sources. We perform prediction experiments using different sets of exogenous variables are compared to make probabilistic forecasts of wind power generation with deep learning models. The exogenous variables considered are the NWP (wind speed + wind direction) and measured weather (wind speed + wind direction) from instruments installed locally in the wind power plant. The wind power plant studied in this work lies in a region with a complex topography in Arctic weather conditions that potentially have a larger source for failure in the NWPs compared to other regions. Therefore it is a motivation in this work to use deep learning models that are trained to account for possible failure sources in weather predictions by using the measured weather to correct for such failures. In the end, we compare the accuracy of the probabilistic forecasts for different sets of covariates and discover the variables that should be included to compute high-quality PIs. 

This paper is structured as follows. In Section 2, an introduction to the field of probabilistic forecasts is provided. Section 3 presents a review of relevant research in the field of probabilistic forecast within energy analytics. In section 4, the wind power plant case study is presented. Section 5 presents the methodology for making probabilistic forecasts. In Section 6 the results are presented in terms of the obtained quality of the PIs. Conclusions are given in Section 7.

\section{The concept of prediction intervals}

Although the field of probabilistic forecasts is based on well-known concepts in statistics, and there exists a vast amount of existing research on this field, some concepts have been shown to be confused and misused \cite{hong2016probabilistic}. To avoid this, we start by defining and explaining some relevant concepts that will be used through the study. 


When making probabilistic forecasts, the goal is to make a PI that considers the uncertainties in the predictions. The PI is an estimate of an interval in which the future observation will fall with a certain probability. For instance, for a 95\% PI, there should be a 95\% probability that the next value will fall within the lower and upper bounds. 

The generic form for computing the upper and lower bound of PIs is
\begin{equation}
    P(Y_n \in C(X_n) ) \geq  1-\alpha,
\end{equation}
where $Y_n$ is the response variable, $C(X_n)$ is the confidence interval centered on the covariate $X_n$, and $\alpha$ is the significance level. For a 95\% PI, $\alpha$ is 0.05.
  
PIs can either be of low quality, or of high quality. A high-quality PI will be sharp and well-calibrated (the PI contains the desired proportion of the samples $1-\alpha$). A low-quality PI is too wide or is miscalibrated (the empirical quantiles do not match the theoretical ones).

\begin{figure}[!ht]
    \centering
    \includegraphics[width=0.6\columnwidth]{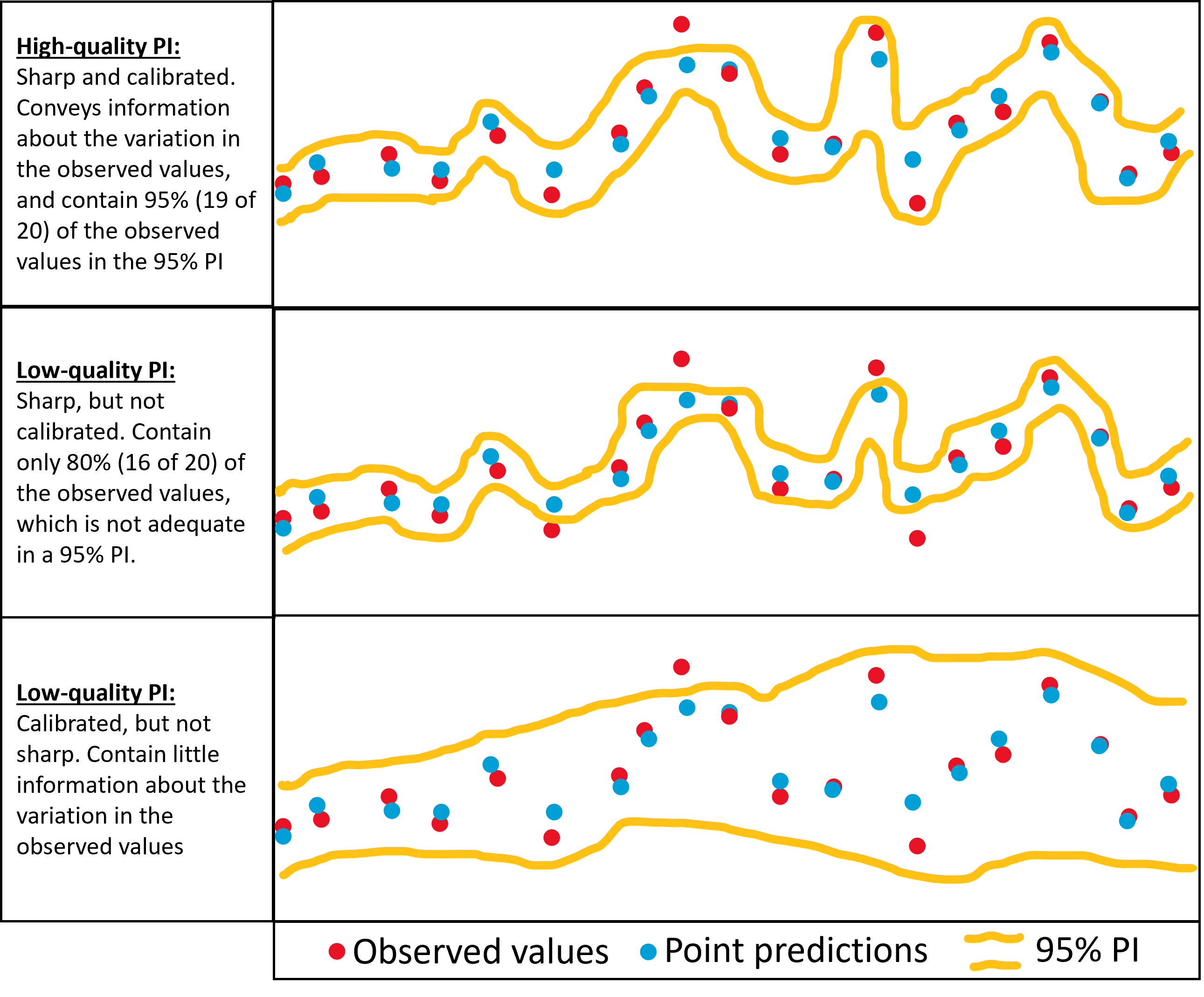}
    \caption{Examples of PI with high (uppermost figure) and low quality. The red dots represent the observed values, the blue dots is the point predictions. The yellow line represent the PI, which in this case is the 95\% prediction interval. }
    \label{fig:PI_eks}
\end{figure}

Examples of high-quality and low-quality PIs is given in Fig.~\ref{fig:PI_eks}. 

Here, we see an example of a point-and probabilistic forecast. The red dots represent the observed values, the blue dots are the point predictions. The yellow line represents the PI, which in this case is the 95\% prediction interval. 
Here, it is clear that the high-quality PI (uppermost Figure) fulfills two important criteria when computing PIs. Sharpness and calibration. The boundaries of the PI should be sharp to convey useful information about the uncertainity in the predictions. In addition, the PI should be calibrated, which means that for a 95\% PI, 95\% of the observed values should fall within the PI. Here, the PI contains exactly 95\% of the observed values, and hence we can trust that there is a 95\% probability that the observed values will fall within this PI. 

The middle and lowermost examples on contrary, show examples of low-quality PIs. The example in the middle shows a PI with high sharpness, but it is not calibrated as it only contains 80\% of the observed values, which is not adequate in a 95\% PI. The lowermost exampled shows a calibrated PI as it contains 95\% of the observed values. However, it is not sharp, and PIs that are too wide do not convey any useful information about the variation in the observed values.

\section{Related works on probabilistic forecasting with deep learning}
\label{sec:relwork}

There exists a vast amount of former research on probabilistic forecast in different applications, and it is outside the scope to review all of them in this work. Therefore, former works that have particularly focused on the application of probabilistic forecast with deep learning within energy applications are reviewed here. In the following, a few examples of some former popular reviews that have high relevance for energy applications is provided.\\ 
In \cite{hong2016probabilistic}, a comprehensive review on probabilistic forecast of the electric load was conducted. The authors offered a tutorial review and presented notable techniques, methodologies, and evaluation methods that could be valuable for researchers and practitioners in the area of load forecasting. To improve the field of probabilistic forecasts, they underlined the need to invest in additional research, and they state the importance of considering emerging technologies and energy policies in the probabilistic load forecasting process. \\
Another relevant review was performed by \cite{zhang2014review}. The authors discuss the challenge of making accurate predictions of the highly intermittent electricity generation from wind power. 
They discuss the value of probabilistic forecasts in such tasks as it could provide additional quantitative information on the uncertainty. For actors that work with decision-making in an uncertain environment (for instance power traders), information about uncertainties in decisions is of interest. 
In this work, the authors presented a review of state-of-the-art methods and new developments within wind power probabilistic forecasting. They discuss different forecast methods and classified them into different categories in terms of uncertainty representation. Finally, they summarized requirements and the overall framework of uncertainty forecasting evaluation. \\

Notably, both these two famous reviews are quite mature (2016 and 2014), and the field of probabilistic forecasting with deep learning in energy analytics has vastly developed in recent years. 
A recent review and comparison study was performed by Mashlakov et al. \cite{mashlakov2021assessing}. Here, the recent developments in the field of probabilistic forecasting, multivariate models, and multi-horizon time series forecasting were reviewed. The authors empirically evaluate the performance of novel deep learning models for predicting wind power, solar power, electricity load, and wholesale electricity price for intraday and day-ahead time horizons. They reviewed the performances of both point and probabilistic forecast approaches. This comprehensive comparison work could serve as a reference point for the quantitative evaluation of deep learning models for probabilistic multivariate energy forecasting in power systems.\\

In addition to the above-mentioned reviews, Tab.~\ref{tab:allworks} present an overview of some recent research works which have contributed to developing the field of probabilistic forecasting with deep learning in energy applications.

\begin{table}[]
\centering
\footnotesize
\setlength\tabcolsep{.9em} 
\caption{Overview of different works}
\label{tab:allworks}
\begin{tabular}{llllll}
\textbf{Model}                     & \textbf{Year} & \textbf{Network} & \textbf{Forecasts} & \textbf{Metrics}                                                & \textbf{Dataset}                                                       \\ \hline
\textbf{CNNs}                      &               &                  &                    &                                                                 &                                                                        \\ \hline
\cite{wang2017deterministic}                           & 2019          & CNN              & Point, Prob.       & \begin{tabular}[c]{@{}l@{}}MAPE,RMSE\\ QL\end{tabular}          & PV power                                                               \\
\cite{chen2020probabilistic}                          & 2020          & TCN              & Prob, Point               &  QL, NRMSE                                                               &   Retailers                                                                     \\
\cite{huang2020improved}                           & 2020          & CNN              & Prob.              & \begin{tabular}[c]{@{}l@{}}RMSE,MAPE\\ PCIP, PINAW\end{tabular} & PV power                                                               \\
\textbf{RNNs}                      &               &                  &                    &                                                                 &                                                                        \\ \hline
\cite{gasthaus2019probabilistic}                           & 2019          & RNN              & Prob               & QL,NRMSE                                                        & \begin{tabular}[c]{@{}l@{}}demand,traffic\\ finance\end{tabular}       \\
\cite{salinas2020deepar}                           & 2020          & RNN              & Prob.              & QL, NRMSE                                                       & \begin{tabular}[c]{@{}l@{}}demand, traffic,\\ electricity\end{tabular} \\
\cite{nguyen2021temporal}                           & 2021          & RNN              & Prob               & CRPS, MSE                                                       & traffic, energy                                                        \\
\cite{alexandrov2020gluonts}                            & 2020          & NN library              & Prob               & QL                                                              & energy, PV                                                             \\
\textbf{Modified NNs} &               &                  &                    &                                                                 &                                                                        \\ \hline
\cite{toubeau2018deep}                           & 2018          & Bi-LSTM          & Prob               & RMSE,QL                                                         & Power market                                                           \\
\cite{xiang2020deterministic}                           & 2020          & BiGRU            & Point, Prob.       & RMSE, QL                                                                & Wind speed                                                                       \\
\cite{zhang2020improved}                           & 2020          & IDMDN            & Prob               & NRMSE                                                           &   Wind power                                                                     \\
\cite{afrasiabi2020advanced}                            & 2020          & GRU+CNN              & Point, Prob                & RMSE, CRPS                                                                &  Wind speed                                                                      \\
\cite{zhang2018improved}                           & 2018          & iQRNN            & Prob.              & QL,WS                                                           & Energy load     \\
\cite{wang2021wind} & 2021 & PCFM & Point, Prob & \begin{tabular}[c]{@{}l@{}}RMSE, MAE\\ PINAW, PCIP \end{tabular} &   Wind speed    \\
\cite{liu2019novel} & 2019 & WT + CNN  &Point, Prob & \begin{tabular}[c]{@{}l@{}}RMSE, MAE, MAPE\\ PINAW, PCIP, CWC \end{tabular}  & Wind speed    \\  
\cite{zhao2018one} &2018 & NARX& Point, Prob &\begin{tabular}[c]{@{}l@{}}RMSE, MAE\\ PINAW, PCIP \end{tabular}  & Wind speed    \\
\cite{zhu2019gaussian} &2019 & LSTM + GPR & Point, Prob & RMSE, MAE, CRPS & Wind speed                                                                                        \\
\textbf{Bayesian methods}          &               &                  &                    &                                                                 &                                                                        \\ \hline
\cite{brusaferri2019bayesian}                            & 2019          & Bayesian         & Prob.              & \begin{tabular}[c]{@{}l@{}}RMSE,MAPE\\ CRPS\end{tabular}        & Energy prices                                                          \\
\cite{liu2020probabilistic}                           & 2020          & STNN             & Point, Prob.              &  RMSE, CRPS                                                               & Wind speed                                                                       \\
\cite{al2021probabilistic}                            & 2020          & Bayesian NNs     &    Point, Prob                &    RMSE, CPRS                                                            &  Energy load                                                                      \\
\cite{kuleshov2018accurate}                           & 2019          & Calibrated       & Prob.              & MAPE, QL                                                        & UCI datasets                                                           \\
\textbf{Ensemble methods}          &               &                  &                    &                                                                 &                                                                        \\ \hline
\cite{wang2017deep}                          & 2017          & WT + CNN         & Prob               & RMSE, PICP                                                      & Wind power                                                             \\
\cite{ahmed2016ensemble}                           & 2016          & 7 ML models      & Prob               & \begin{tabular}[c]{@{}l@{}}RMSE,MAE\\ QL\end{tabular}           & Solar power                                                            \\ \cite{jin2021probabilistic}     & 2021  &  SEFMGPR & Prob & RMSE, R$^2$&   Wind power                                                                           \\
\cite{la2020added} & 2020 & EPS & Prob & CRPS & Solar irradiance \\ 
\cite{sun2020multi} & 2020 & MDE & Prob & QL & Wind power \\
\hline
\end{tabular}
\end{table}

All works in Tab.~\ref{tab:allworks} have tested the performance of their proposed methodology on energy-related datasets, and the prediction performances are presented with popular probabilistic metrics such as the QL, CRPS, PCIP, and PINAW (some of these metrics are defined and used in the result section in this work). The methodologies involve either using CNNs, RNNs, or some modifications of them. In addition, some work has proposed methodologies where they ensemble multiple NNs together to make probabilistic forecasts. Most works have applied the frequentist approach where the main aim is to optimize a cost function (such as the quantile loss), but some works have also proposed a Bayesian approach to make probabilistic forecasts with deep learning. Noteworthy, among all works, there is no specific model or method that seems to consistently achieve the best results, apart from the DeepAR model which has shown promising results in several works \cite{mashlakov2021assessing,salinas2020deepar,alexandrov2020gluonts}. 

In a recent review and comparison work by \cite{mashlakov2021assessing}, the authors highlighted the need of testing more automated deep learning models to progress the research within the field of probabilistic forecasts in energy applications. The authors suggested using open source libraries for modeling, such as the GluonTS toolkit \cite{alexandrov2020gluonts}. The aim of the GluonTS library developed by \cite{alexandrov2020gluonts} is to provide a flexible tool for probabilistic time series modeling with deep learning-based models. 
Motivated by the suggestion in \cite{mashlakov2021assessing}, we implement deep learning models from the GluonTS library to make probabilistic forecasts of wind power generation. 

\section{Wind power plant case study}
\subsection{The wind power plant located in Northern Norway}
\begin{figure}[!ht]
    \centering
    \includegraphics[scale=0.75]{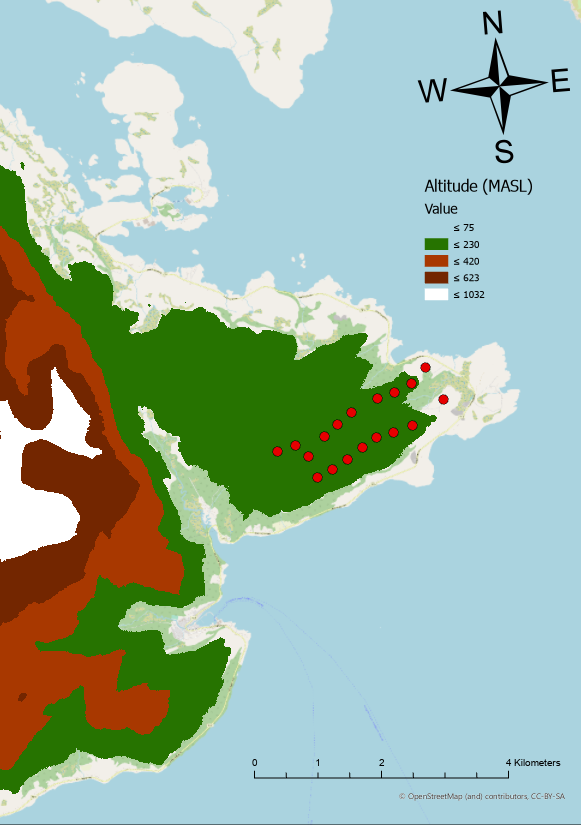}
    \caption{Altitude map where the different colors indicate the altitude level. Green color is altitudes between 75 and 230 MASL. The highest mountains around in the region close to the wind park is up to 1032 MASL. The 18 turbines in the wind farm are marked in red dots.}
    \label{fig:Fakken}
\end{figure}

In this work, the power generation from a wind farm in the region of Northern Norway is predicted. This wind farm consists of 18 turbines with a maximum power generation capacity of 3 MW, giving a maximum capacity of 54 MW for this power plant. In Fig.~\ref{fig:Fakken}, an altitude map is created for the region of the wind power plant. 

The different colors in the altitude map indicate the interval in Meter Above Sea Level (MASL). Here, the green color is all altitudes between 75 and 230 MASL. The highest elevation in the mountains to the right has an altitude between 632 and 1,032 MASL.
The red dots are the geographical positions of the different turbines. Most of the turbines are approximately at the same altitude below 230 MASL. The two turbines at the rightmost positions are slightly below an altitude of 75 MASL. 

Such a complex topography where the altitude varies from 0 to above 1000 MASL within small regions, is a typical phenomenon for the terrain in Northern Norway. This terrain has a huge impact on the local weather variation, and this particular wind power plant is highly affected by local variations in the wind. The owner of this wind power plant reported that the leftmost turbine produced 25\% less energy than the rightmost turbine during the year 2020. The reason is that the turbines located to the left are shielded by the large mountains (indicated by the white color in the altitude map) which reduce the amount of wind that hits the turbine. On the other hand, the turbines located to the right are close to the ocean where there are few objectives that could reduce the wind. This power plant is a perfect example of where it can be large differences in production between the turbines. This can increase the difficulties in making predictions for the whole park, as the total power output from the park will be even more fluctuating compared to wind farms in flat regions where the weather conditions are more equal throughout the whole farm. 


\subsection{Dataset and and technical properties of wind turbines}
\begin{figure}[!ht]
    \centering
    \includegraphics[width=0.5\columnwidth]{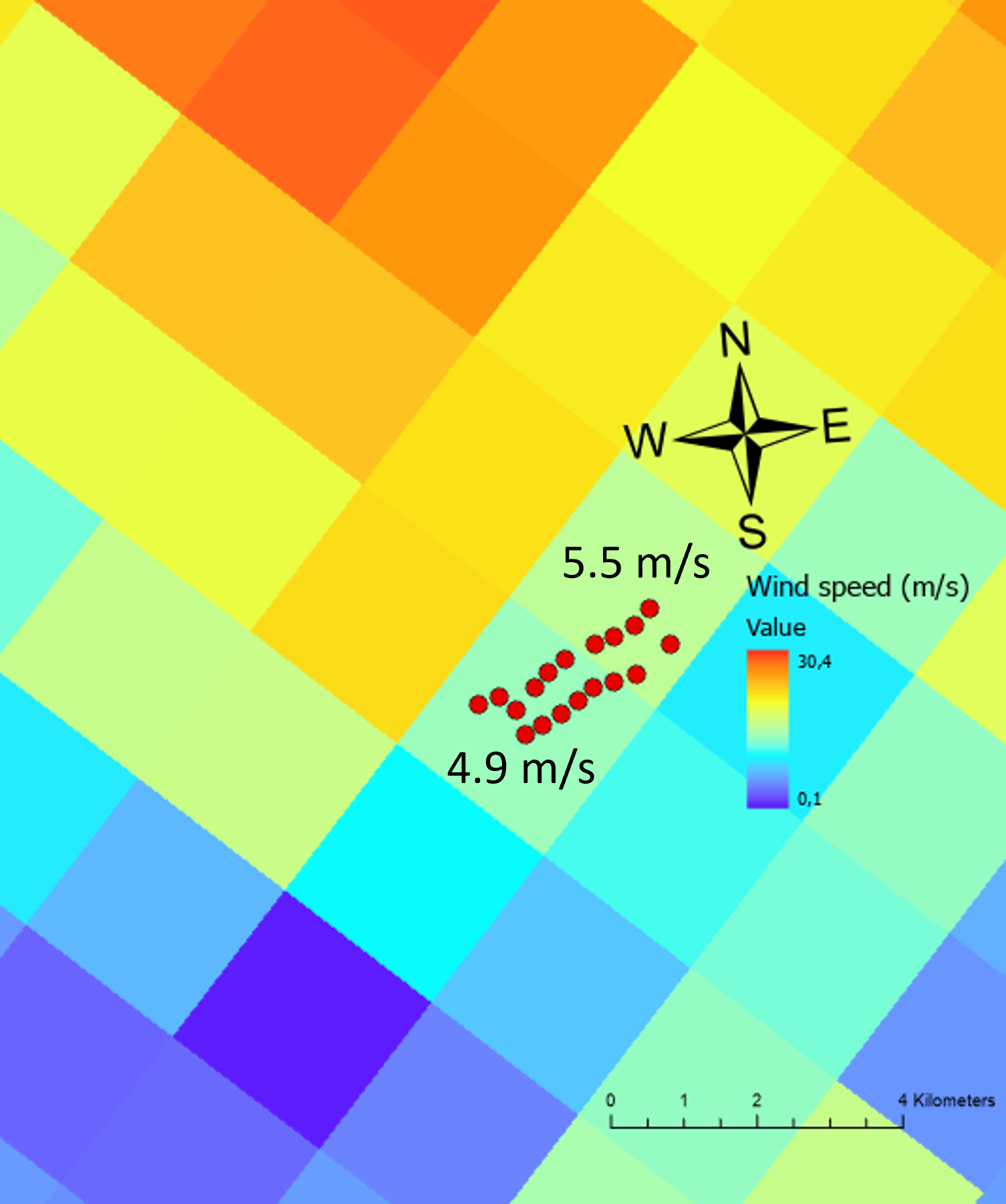}
    \caption{The Arome Arctic weather simulation map where the red dots represent the position of each wind power turbine. Each square represent the spatial resolution of 2.5$\times$2.5 kilometers. 12 turbines and 6 turbines are located in the leftmost and rightmost cell, respectively}
    \label{fig:AromeArctic}
\end{figure}

The data available from the wind park is the historical measured power generation, wind speed, and wind direction measured on each turbine from the year 2020. The measured wind speed and wind direction variables are collected from the weather stations mounted on the nacelle on the turbine that is located in the middle of the wind park. The data are in a 1-hour resolution and are received from the Troms Kraft Power company, the owner of the wind power plant. This gives a dataset of totally 8,784 samples.
In addition to measured data, the NWP from the AROME-Arctic model is collected. This model is developed by the meteorological institute of Norway (MET) and provides weather forecasts with a spatial resolution of 2.5 kilometers and a temporal resolution of 1 hour. Similar to the weather measurements, the predicted wind speed and wind direction variables are collected from the AROME-Arctic model. Using only wind speed and wind direction as additional input variables to the prediction experiments has been shown to provide the most accurate forecasts in former wind power forecast literature \cite{monteiro2009wind}. This is due to the fact that wind power generation is directly affected by the amount of wind that hits the turbine blades. Hereafter, the weather data is referred to the wind speed and wind direction. 
Fig.~\ref{fig:AromeArctic} illustrate a map with the AROME-Arctic simulated wind speed for a randomly selected hour on the 1st of March in 2020. The colors in the map represent the simulated wind speed in each cell (spatial resolution of 2.5 kilometers). It is clear that the turbines in the wind power park are distributed over two different cells, and there are differences in the wind speed for the two cells.

Fig.~\ref{fig:AromeArctic} show large differences in wind speed within a few kilometers, ranging from zero wind (blue color in southernmost part) to almost 30 m/s wind (red color in northernmost part). The AROME-Arctic weather forecast for this particular example shows that the rightmost part of the wind power plant has a wind of 5.5 m/s, and the leftmost part has a wind of 4.9 m/s. To consider possible local differences in wind speed and wind direction, the weather data from both cells in the AROME-Arctic weather simulation map is included as exogenous variables.  

\begin{figure}[!ht]
    \centering
    \includegraphics[width=0.8\columnwidth]{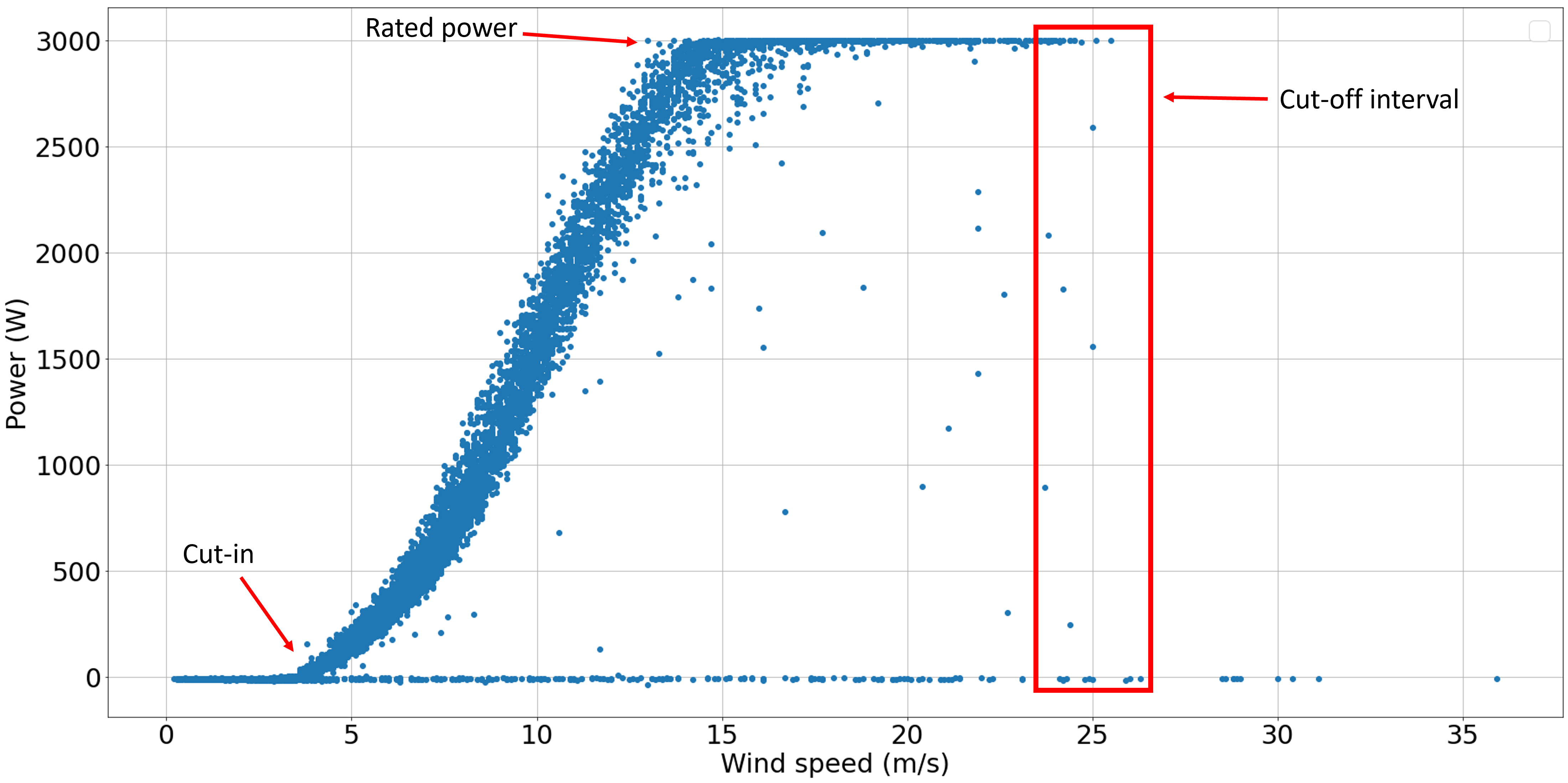}
    \caption{Power curve from a wind turbine in the wind farm studied in this work with the cut-in, rated and cut-off wind speed marked. The zero production above 4 and below 25 m/s wind represent periods where the turbine is shut down due to turbine failure or maintenance}
    \label{fig:PowerCurve}
\end{figure}

Before proceeding to the prediction methodology, some technical features of wind power generation must be highlighted. Wind power generation is a (RES) technology that has a highly intermittent power generation due to the strong dependency on weather conditions. In addition, the power generation from the wind turbines is dependent on the \emph{power curve}. The power curve from one of the turbines in the wind farm analyzed here is given in Fig.~\ref{fig:PowerCurve}.

As seen in Fig.~\ref{fig:PowerCurve}, the wind turbine has zero production between 0 m/s to 3/4 m/s. At the cut-in wind speed around 4 m/s, the power generation increases towards a wind speed of 12-13 m/s. At this wind speed, the power production reaches the maximum limit (or the rated power) of 3MW around and produces at maximum it suddenly drops towards zero in the cut-off interval around 25 m/s. The reason for this drop is due to safety. If the blades at the turbine rotate too quickly, it can damage the equipment. Therefore, the turbines are forced to stop.

\section{Methodology}
\subsection{Dataset configuration, preliminary analyses and deep learning models}
In this work, the aim is to predict the day-ahead wind power generation. The day-ahead market closes at 12:00 where the market participants must submit their final bids to the electricity market about the expected amount of power generation the next day. Therefore, in this work, the forecast horizon is 36 hours to consider all hours in the next day (12 hours + 24 hours). 

\begin{figure}[!ht]
    \centering
    \includegraphics[width=\columnwidth]{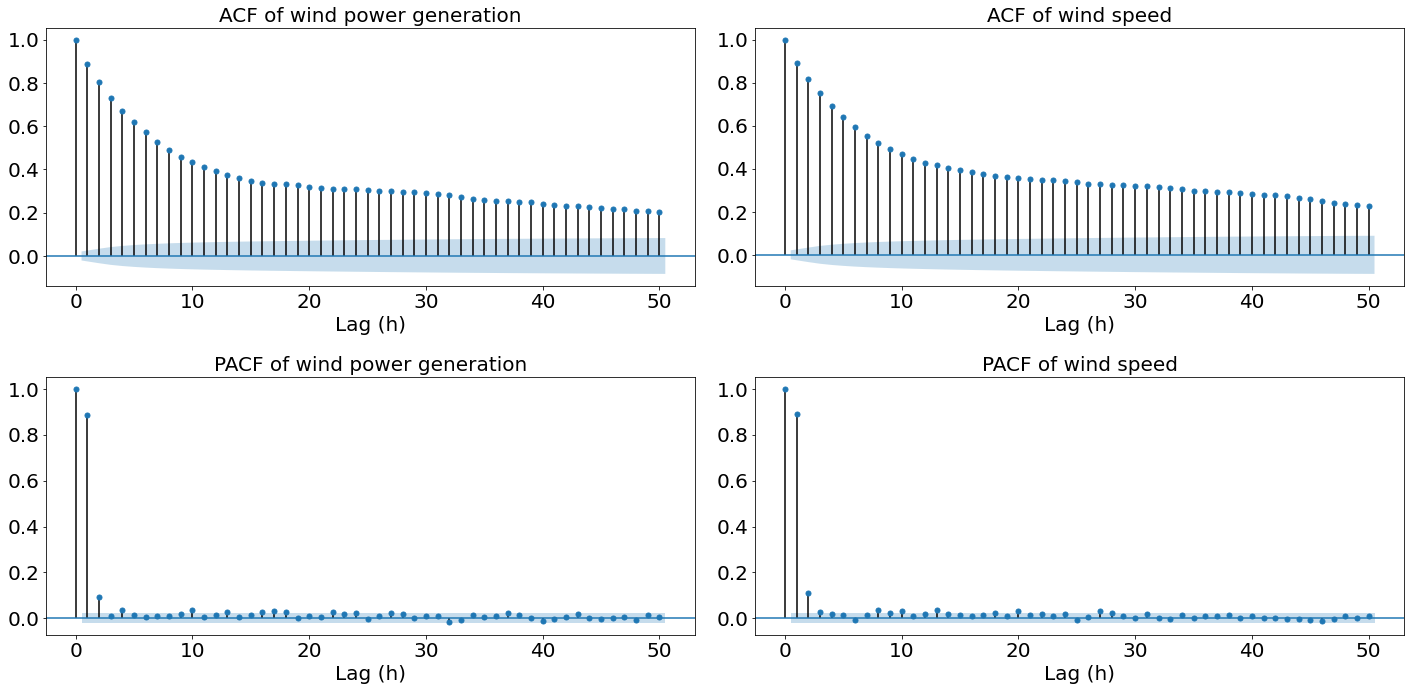}
    \caption{ACF and PACF of the time series on wind power generation and wind speed in the year of 2020. The correlation outside the standard deviations are correlations and not a statistical fluke.}
    \label{fig:ACFandPACF}
\end{figure}

First, in order to identify the time varying patterns in the time series, some statistical analyses are performed by computing the autocorrelation function (ACF) and the partial autocorrelation function (PACF). Fig.~\ref{fig:ACFandPACF} shows the ACF and PACF graphs for the time series of the wind power generation and wind speed in the year of 2020.
The ACF and PACF plots show no repetitive patterns, which indicate lack of seasonality in the time series. Interestingly, the time series show short-term dependencies as the ACF show correlations outside the 95\% confidence interval for all time lags (depicted as a blue area in Fig.~\ref{fig:ACFandPACF}). The values outside of the blue cone are very likely actual correlations. The PACF plots do not show any strong correlations except from the first three lags, and the correlations in the ACF plot after lag 3, are indirect correlations that can be explained by the first three time steps. Noteworthy, the ACF and PACF plots capture only linear dependencies in the wind power generation and wind speed, but there could be non-linear dependencies in the time series in addition. The deep learning models used in this work provide the capability to capture such non-linear relationships. 

The correlation in the ACF and PACF plots shows that historical data on wind power generation should be used as input to the deep learning models to make predictions of the expected power generation. In addition, as wind power generation is directly affected by the amount of wind that hits the turbines, it is of interest to investigate which variables should be included (or excluded) as covariates to potentially improve the prediction performance. The exogenous variables included in addition to the historical data on wind power generation are the historical measurement data on weather conditions and the NWP from the AROME-Arctic model.
The measured weather is included as input due to the complex terrain in the region of the wind farm, and the weather model with a spatial resolution of 2.5 km might be too coarse to capture local variations in the wind. Therefore, the information from measured wind from the nacelle could provide important additional information.

The prediction experiments are summarized as; 1) Use historical measurement data on power generation with historical measurement data on weather, and the 36 hour ahead NWP as exogenous variables to predict the future power generation. 2) Use historical measurement data on power generation, where only measured weather data is used as exogenous variables. 3) Use historical measurement data on power generation, where only the 36-hour ahead NWP data is used as exogenous variables. 
%
\begin{figure}[!ht]
    \centering
    \includegraphics[width=\columnwidth]{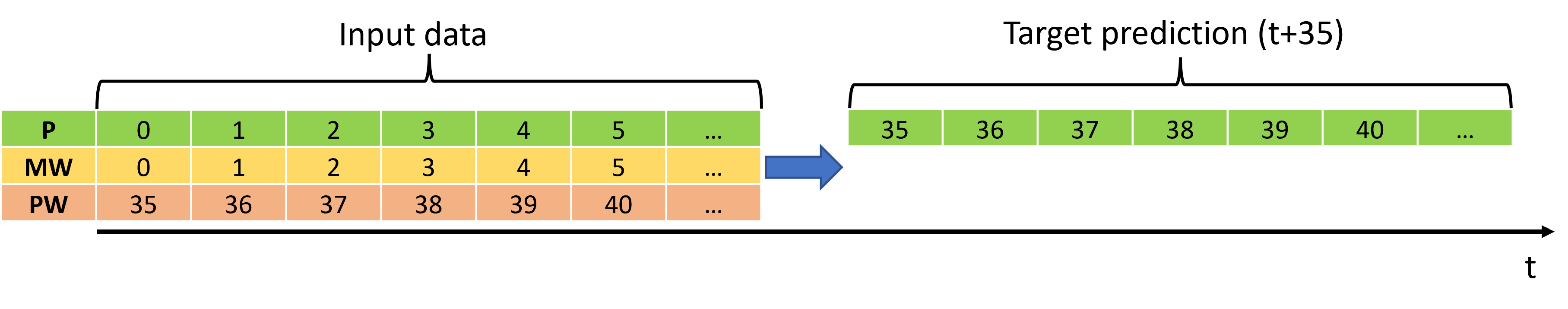}
    \caption{The historical data on wind power generation (P) togheter with historical data on measured weather (MW) and predicted weather (PW) to predict the future wind power generation (target prediction). }
    \label{fig:dataset}
\end{figure}

In Fig.~\ref{fig:dataset} the example on the first prediction experiment where both measured weather (MW) and predicted weather (PW) is used as additional inputs to predict the future power generation is illustrated. In prediction experiments 2) or 3), either PW or MW is removed as input.

To make predictions with the different covariates, the DeepAR model proposed by \cite{salinas2020deepar} is selected as it has outperformed other models in several recent works on probabilistic forecasts with deep learning \cite{mashlakov2021assessing,salinas2020deepar,alexandrov2020gluonts}. 
The inputs to the DeepAR models are the time series values (historical wind power generation) until $t-1$ and potential covariates (such as measured and predicted weather) at time $t$.
The covariates and the time series values are thereafter concatenated before being fed into the internal units, which can either be LSTM or GRU layers. The output from the internal units is fed into two different linear layers. One for computing the mean, and one for computing the standard deviation. When computing the standard deviation, the linear layer is fed into a SoftPlus layer to ensure positive values. In the end, the computed mean and standard deviation are used as input to a Gaussian likelihood model where predictive samples are generated. During training, the model learns by maximizing a Gaussian log-likelihood function and is optimized via stochastic gradient descent with respect to the model parameters. The DeepAR model is described more in detail in \cite{salinas2020deepar}.

In former works the default DeepAR configuration with internal LSTM units has been used \cite{salinas2020deepar,mashlakov2021assessing,alexandrov2020gluonts}. In this work, the DeepAR model is tested with Gated Recurrent Units (GRU)~\cite{bianchi2017recurrent} in addition to LSTM as internal units. 

In this work, two versions of a persistence model are served as benchmarks for the DeepAR model. The persistence model assumes that the wind power generation at a certain future time will be the same as it is when the forecast is made (for instance, if the generation is 50MW when the prediction is made, the persistence model assumes that the generation the next hour also is 50MW). The persistence model has been a popular model to use as a benchmark as it is a simple method to implement, and are often a difficult method to outperform, especially in the range of 1-6 hour ahead \cite{monteiro2009wind}. In addition, a modified version of the persistence model is served as a benchmark model in this work. The authors in \cite{moehrlen2004uncertainty} proposed a modified persistence model for predicting wind power generation. This model is a combination of the mean of the time series, where the future value is weighted as a function of the correlation between the current and average power, respectively. In prediction experiments with a forecast horizon above 3 hours, the modified persistence model has been shown to outperform the original one \cite{moehrlen2004uncertainty}.


\subsection{Prediction strategy}
To train and evaluate the models, the time series is split into a training set (85\%) and a test set (15\%). The training set is further divided into training (80\%) and validation (20\%). The training set is used to fit the model parameters by minimizing the prediction loss, and the validation set is used to find the optimal configuration of the hyperparameters. Each model is set to train on 500 epochs, and to avoid potential overfitting during training, an early stopping rate is introduced. 
A grid search was performed to find the optimal hyperparameter configuration for each model. We selected the configuration that resulted in the highest accuracy on the test set. The search space for the selected hyperparameters and the optimum configuration for each model is presented in Tab.~\ref{tab:HyperSearch}. 

\begin{table}[!ht]
\centering
\footnotesize
\setlength\tabcolsep{.9em} 
\caption{Details of the hyperparameter search space and optimal configurations for different models}
\label{tab:HyperSearch}
\begin{tabular}{llll}
\hline
Parameter             & Search space                                                         & Optimal DeepAR$_{\text{GRU}}$    & Optimal DeepAR$_{\text{LSTM}}$     \\ \hline
Rolling window length & 36, 72                                                               & 36                       & 36                       \\
Layers                & 1, 2, 3                                                              & 2                        & 2                        \\
Hidden units          & 32, 64                                                               & 32                       & 64                       \\
Dropout rate          & 0.0, 0.1, 0.2                                                        & 0.0                      & 0.2                      \\
Learning rate         & $10^{-4}$,$10^{-3}$,$10^{-2}$ & $10^{-2}$ & $10^{-3}$ \\
\hline
\end{tabular}
\end{table}

The searched hyperparameters include the length of the rolling window, which is the number of time-points that the model gets to see before making the prediction, number of hidden units, number of layers, dropout rate, and learning rate. The rest of the hyperparameters were not fitted, and the default values were used. 

\section{Experimental evaluation and results}
\subsection{Evaluation metrics}
When making point predictions, the main aim is to minimize the discrepancy between the predicted and true output, respectively. For the purpose of probabilistic forecasts, it is more complicated. Here one wants to minimize the loss between the predicted and true value for a specific quantile level (or percentile which are the upper and lower bound of the PI), but at the same time, one wants to have a PI that contains the true outcome. In addition, one does not want to have a PI that is too wide as it will contain no useful information, so the PI should be as sharp as possible, but still contain the true values. 
In the following, some popular metrics that will be used in the result section to provide the skill score of both point-and probabilistic forecasts are described.

A widely used metric to evaluate the performance of \emph{point forecasts} is the root mean squared error (RMSE) and the Normalized RMSE (NRMSE). The RMSE is defined as
\begin{equation}
    RMSE = \sqrt{\frac{1}{n} \sum_{i=1}^{n}(\hat{y}_i-y_i)^2},
\end{equation}
where $\hat{y}$ and $y$ are the predicted and true values. The RMSE measures the discrepancy between the predicted values and observed values at time $i$, over $n$ number of observations. The NRMSE relates the RMSE to the observed average value in the observation period and is defined as:
\begin{equation}
    NRMSE = \frac{RMSE}{\overline{y}},
\end{equation}
where $\overline{y}$ is the average of the observed values in the time series.

Another popular method to evaluate the accuracy of point forecasts are the mean absolute percentage error (MAPE). MAPE is defined as:
\begin{equation}
    MAPE = \frac{100}{n}\sum_{i=1}^{n} \vert \frac{y_i - \hat{y}_i}{y_i} \vert,
\end{equation}

The common feature of all these metrics (RMSE, NRMSE, and MAPE) is that the lower value, the higher accuracy. 


The purpose of \emph{probabilistic forecasts} is to have a PI that fulfills the calibration and sharpness criteria. In this section, some common metrics that are used to describe the performance of the PIs in terms of the sharpness and calibration criteria are presented. 

The pinball-loss (PL), or the quantile-loss (QL) function is a common metric to measure the performance of the PI. For each quantile level, the PL function returns a value that can be interpreted as the accuracy of a quantile forecasting model.
Let $q$ be the target quantile, $y$ the real value, and $\hat{y}$ the quantile forecast, then the PL for quantile $q$ can be written as:

\[PL_q (y,\hat{y})= \left\{
  \begin{array}{lr}
    (y-\hat{y})q & y\geq \hat{y}\\
    (\hat{y}-y)(1-q) & \hat{y} > y
  \end{array}
\right.
\]
The PL function penalizes the forecast if the model is over or under-predicting depending on the quantile level that is computed. The large quantile level will be more penalized for under-predicting than a low quantile level. Similarly, a low quantile level will be more penalized for over-predicting than a large quantile level. This makes sense as in the high quantile level case, one expects most of the observed values to be smaller than the predictions, and at the low quantile level one expects most of the values to be above the predicted values. 
Similar to the point forecast metrics, the lower PL, the more accurate the quantile forecast is. If computing the PL over a set of different quantiles levels, the final quantile loss result will be the average of all levels (often denoted as QLm).

In \cite{salinas2020deepar}, the authors evluated the sharpness, or the quantile risk for different quantile levels by considering the normalized sum, wQL of quantile losses. The wQL for a quantile $q$ is computed as all pinball losses divided by the sum of true output:
\begin{equation}
    wQL_{q} (y,\hat{y}) = 2 \frac{\sum_{i} PL_{q} (y_{i},\hat{y}_{i})}{\sum_{i}y_i}
\end{equation}
A low wQL indicate a sharper PI.





Besides sharpness, it is important to compute how calibrated the PI is. A popular calibration metric for probabilistic forecasts is the Prediction Interval Coverage Probability (PICP) \cite{huang2020improved}. The PICP is employed to compute the probability that the true outcome is within the PIs. The PCIP is defined as
\begin{equation}
    PCIP = \frac{1}{n}\sum_{i=1}^{n}u_i,
\end{equation}
where $n$ is the total number of samples. When the true output is within the upper and lower bound, $u_i=1$, otherwise $u_i=0$. To obtain a well-calibrated PI, the coverage should be close as possible to the PI that is specified. For instance, for a 95\% PI, the PCIP should be 0.95. 

The final metric that is used to measure the quality of the PIs is the Mean Scaled Interval Score (MSIS). This metric was used as the preferred one in the M4 forecasting competition, where 100,000 time series and 61 forecasting methods were compared \cite{makridakis2020m4}. The MSIS score was proposed by \cite{gneiting2007strictly} and evaluate the performances of the generated PIs as
\begin{equation}
    MSIS = \frac{1}{h} \times \frac{\sum_{t=n+1}^{n+h}(U_t - L_t)+\frac{2}{\alpha}(L_t - Y_t)  \mathbbm{1} Y_t < L_t + \frac{2}{\alpha}(Y_t - U_t)  \mathbbm{1} Y_t > U_t}{\frac{1}{n-m}\sum_{t=m+1}^{n} \vert Y - Y_{t-m} \vert },
\end{equation}
where the lower and upper bounds of the PI are denoted by $L_t$ and $U_t$, respectively. $Y_t$ is the future observed values, $h$ is the forecast horizon, and $\mathbbm{1}$ is the indicator function (1 if $Y_t$ is within the PI and 0 otherwise). Here $\alpha$ is the significance level, and for a 95\% PI, $\alpha$ is set to 0.05.
The MSIS metric deals with the sharpness and calibration criteria. It both penalize wide PIs (since $U_t$ and $L_t$ will be large), and penalize non-coverage.  
Here, a lower MSIS score indicates a better PI in terms of sharpness and calibration.

\subsection{Results and discussion}
In Tab.~\ref{tab:Gluonresults} the results are given. In this work, the 95\% PI is computed as it is a widely used choice for economic, financial, and energy-related forecasting applications \cite{makridakis2020m4}. The point prediction scores for predicting the 36-hour ahead power generation is given in terms of NRMSE and MAPE (eq. (7) and (8)). For probabilistic forecasts, the scores are given in terms of MSIS, PCIP, and the wQL (eq. (9)-(11)). In Tab.~\ref{tab:Gluonresults}, the mean wQL for quantile level 0.025 and 0.975 is computed. 


\begin{table}[!ht]
\centering
\footnotesize
\setlength\tabcolsep{.9em} 
\caption{36-hour ahead prediction scores with different dataset configurations}
\label{tab:Gluonresults}
\begin{tabular}{lllllll}
\multicolumn{7}{c}{Configuration   1: Measured and predicted weather (wind speed + wind direction)}                                           \\ \hline
Model        & NRMSE         & MAPE          & MSIS          & PCIP$_{2.5}$      & PCIP$_{97.5}$     & mean wQL       \\ \hline
DeepAR$_{\text{GRU}}$  & 0.17 & 0.16 & 3.56 & 0.00 & 1.00 & 0.028\\
DeepAR$_{\text{LSTM}}$ & \textbf{0.16}          & \textbf{0.15}          & \textbf{2.53}            & \textbf{0.027}            & \textbf{0.972}          & \textbf{0.020}          \\
             &               &               &               &                &                &                \\
\multicolumn{7}{c}{Configuration 2: Measured weather (wind speed + wind direction)}                                                               \\ \hline
DeepAR$_{\text{GRU}}$ & 0.45          & 0.41         & 10.73          & 0.00          & 0.55           & 0.084          \\
DeepAR$_{\text{LSTM}}$ & 0.40          & 0.35          & 10.16          & 0.00           & 0.55           & 0.080          \\
             &               &               &               &                &                &                \\
\multicolumn{7}{c}{Configuration 3: Predicted weather (wind speed + wind direction)}                                                              \\ \hline
DeepAR$_{\text{GRU}}$  & 0.24          & 0.21         & 5.32          & 0.00              & 1.00              & 0.042          \\
DeepAR$_{\text{LSTM}}$ & 0.29          & 0.25          & 4.03         & 0.00           & \textbf{0.972}           & 0.032    \\
\multicolumn{7}{c}{Benchmark models}                                                              \\ \hline
Persistence & 1.06 & 6.17 & - &- &- &- \\
Modified Persistence & 0.79 & 5.09 &- &- & -&-\\

\end{tabular}
\end{table}

Tab.~\ref{tab:Gluonresults} show that the DeepAR model with LSTM units resulted in the best accuracy for all experiments. The Persistence models obtained the worst prediction performance. The probabilistic scores are not reported for the persistence models as they do not provide the capability to make probabilistic forecasts. 

The best performance was obtained with the dataset configuration where both measured and predicted weather was included as exogenous variables.
With this configuration, the DeepAR$_{\text{LSTM}}$ model computed the sharpest PI as the mean wQL and the MSIS are low. In addition, this model computed a PI that is well-calibrated as the PCIP is close to the quantile levels that are specified. For instance, for PCIP$_{2.5}$, the computed PI resulted in a PCIP of 2.7\%, which indicates that it is a probability of 2.7\% that the true outcome is below this quantile level. The PCIP for the upper boundary is also close to the specified quantile level. 

Using only measured weather as exogenous variables resulted in the worst prediction performance for both models. The computed PI for this configuration is neither sharp nor calibrated, as the MSIS and wQL are large and the PCIP$_{97.5}$ value is 0.55. This shows that the true outcome is below the upper PI boundaries only 55\% of the time, which is not adequate in a 95\% PI interval. 

When the NWP is used as the only exogenous variable, acceptable prediction performances were obtained for the day-ahead predictions. However, indicated by the MSIS and wQL scores, the PI is less sharp in this case. In addition, the PI is not fully calibrated for both models as there is zero probability that the true outcome is below the lower PI boundary.

The results indicate that combining historical measurement data and NWPs helps improve the day-ahead prediction of wind power generation. 
This shows that adding historical data on measured weather allows the DeepAR model to auto-correct systematic biases in the NWPs.



\begin{figure}[!ht]
    \centering
    \includegraphics[width=0.8\columnwidth]{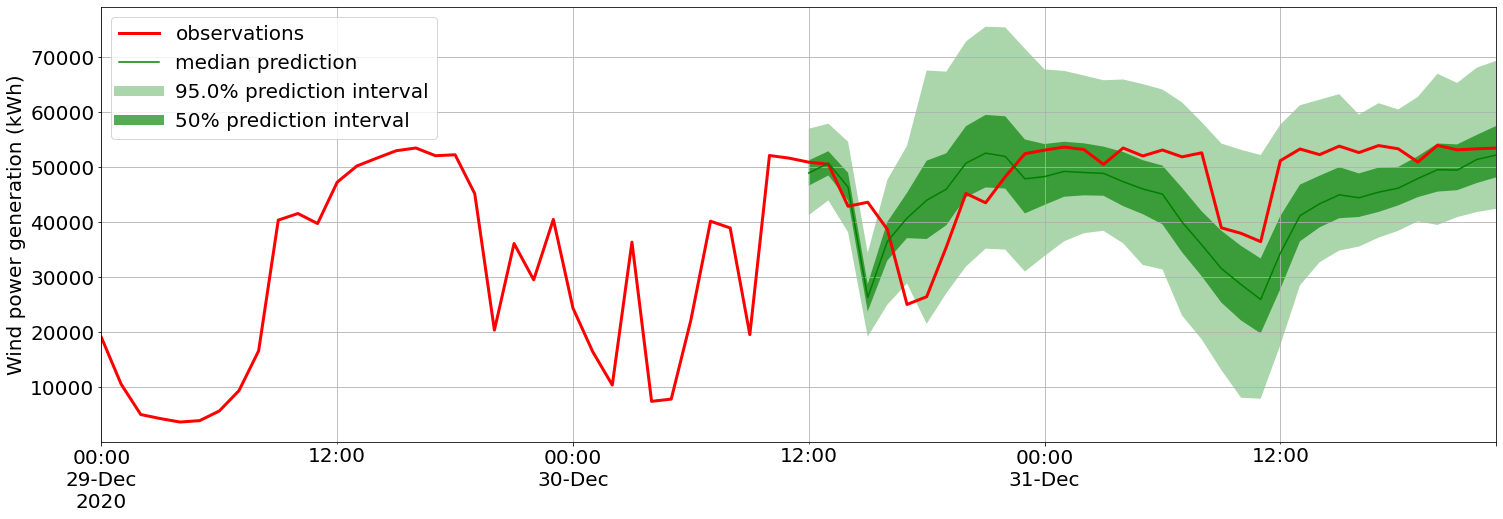}
    \includegraphics[width=0.8\columnwidth]{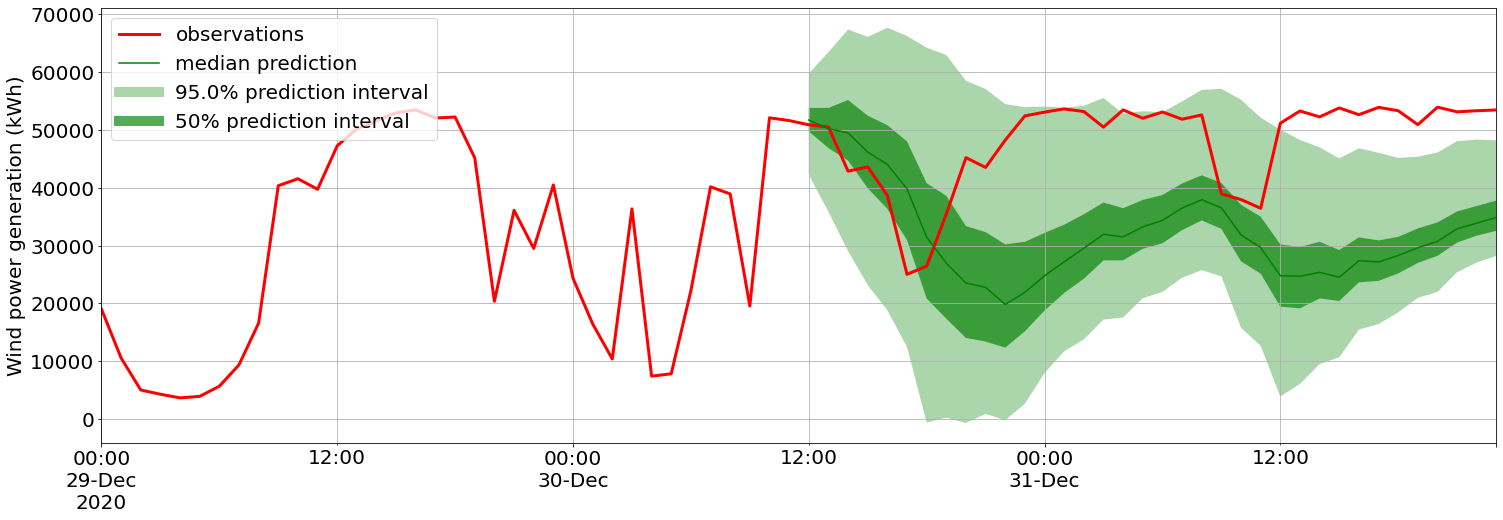}
    \includegraphics[width=0.8\columnwidth]{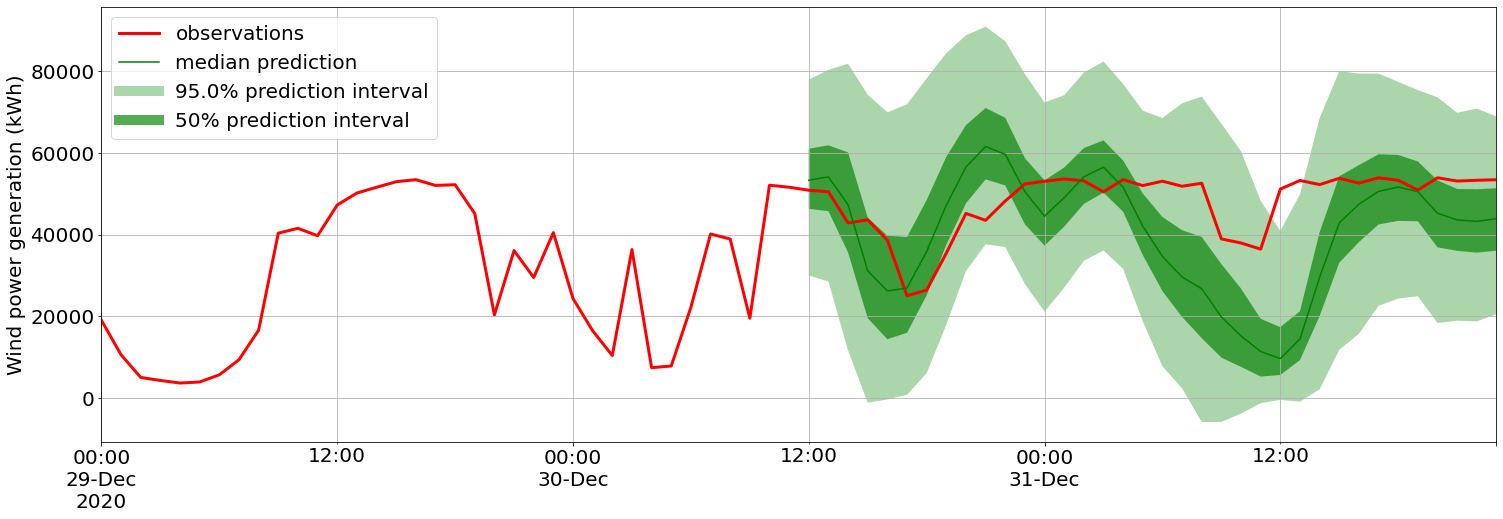}
    \caption{36 hour ahead predictions of wind power generation. The uppermost Figure show the DeepAR$_{\text{LSTM}}$ model where the measured wind and NWP are included as exogenous variables. The Figure in the middle shows the DeepAR$_{\text{LSTM}}$ model for configuration 2 where only measured weather is used. This shows a low degree of sharpness and are miscalibrated. The lowermost Figure shows the result from the DeepAR$_{\text{LSTM}}$ model where only the predicted wind is used as exogenous variables. This result show a well calibrated PI, but low degree of sharpness as the PI boundaries is wide compared to the first graph, and provide less useful information about the variation in expected output.}
    \label{fig:DeepAR_results}
\end{figure}

In Fig.~\ref{fig:DeepAR_results} the 36-hour ahead predictions with the best DeepAR$_{\text{LSTM}}$ model in terms of calibration and sharpness is provided (configuration 1 with measurements and NWPs combined). In addition, the DeepAR$_{\text{LSTM}}$ for configuration 2 and 3 is shown. These are two examples of PI results that are miscalibrated and have low sharpness. In the following illustration, the 50\% PI and the median prediction are included. 

The red line represents observed values, while the green line represents the median value (or point forecast) of the probabilistic forecast. It is seldom a perfect match between the predicted point forecast and the actual value. This is not surprising as making accurate point forecasts of the day-ahead wind power generation is a very difficult problem. The green nuances in the graphs represent the different PIs. The uppermost graph in Fig.~\ref{fig:DeepAR_results} show the resulting prediction with the DeepAR$_{\text{LSTM}}$ model using configuration 1. In this prediction, the 95\% PI indicated by the bright green color shows that the actual measurements fall within the interval approximately 95\% of the time and therefore show a well-calibrated PI. The observed values fall outside the interval at some points in the period between 12:00 and 18:00 the 30-Dec. This is acceptable, as given by the 95\% requirement, some values may fall outside the range. The rest of the time the observed values are within the 95\% PI. The graph also shows that the model is more confident at the beginning of the prediction period as the PI boundaries are very sharp, but the uncertainties increase as longer ahead in the future the predictions are made. 

On contrary, the graph in the middle shows the DeepAR$_{\text{LSTM}}$ model for the configuration using only measured wind. This shows a low degree of sharpness and is not calibrated as several observations are outside the PI boundaries. The lowermost graph shows the result from the DeepAR$_{\text{LSTM}}$ model where only the predicted wind is used as exogenous variables. This result shows a well-calibrated PI as most observations are within the boundaries, but the PI is less informative as is it wide and consequently provides less useful information about the variations in output. 

\begin{figure}[!ht]
    \centering
    \includegraphics[width=0.9\columnwidth]{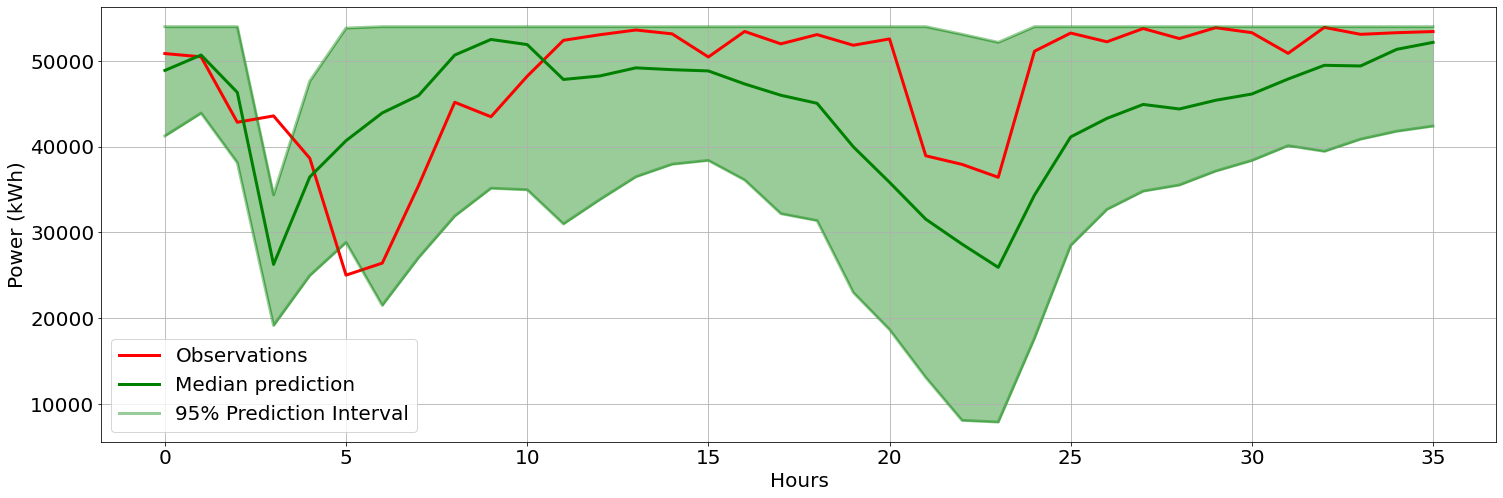}
    \caption{36 hour ahead probabilistic forecast where the upper boundaries of PI are adjusted with respect to the theoretical maximum power generation of 54 MW.}
    \label{fig:Modified_PI}
\end{figure}

Due to technical limitations, the wind power plant can maximum produce 54 MW (18 turbines $\times$3 MW). However, all models in Fig.~\ref{fig:DeepAR_results} compute a PI that has upper boundaries that exceed this value. This is not possible due to the theoretical maximum of 54 MW, and one can be sure that there is zero probability that the total wind power generation will exceed this value. Therefore, all values above 54 MW in the upper boundary of the computed 95\% PIs can be replaced by the theoretical maximum power generation. In Fig.~\ref{fig:Modified_PI} the modified 95\% PI of the DeepAR$_{\text{LSTM}}$ with the best prediction performance is provided. Here all values above the theoretical maximum are replaced by the theoretical maximum of 54MW.

Fig.~\ref{fig:Modified_PI} show that replacing the upper PI boundaries with the theoretical maximum power output from the wind farm generates a PI that is much sharper and is very accurate at several time steps. This indicates that when making a probabilistic forecast of power output from wind farms, knowledge regarding technical limitations could contribute to achieving even more accurate PIs. The scores of the original and adjusted 95\% PI for the DeepAR$_{\text{LSTM}}$ for configuration 1 is given in Tab.~\ref{tab:AdjustedPI}. 

\begin{table}[!ht]
\centering
\footnotesize
\setlength\tabcolsep{.9em} 
\caption{Probabilistic performance scores with the original and adjusted 95\% PI}
\label{tab:AdjustedPI}
\begin{tabular}{lclll}
\multicolumn{1}{c}{} & \multicolumn{2}{c}{Original 95\% PI}                     & \multicolumn{2}{l}{Adjusted 95\% PI}                 \\ \hline
Model                & \multicolumn{1}{l}{MSIS} & \multicolumn{1}{l|}{mean wQL} & MSIS                     & mean wQL                  \\ \hline
DeepAR$_{\text{LSTM}}$        & 2.53                     & \multicolumn{1}{c|}{0.020}    & \multicolumn{1}{c}{1.16} & \multicolumn{1}{c}{0.013}
\end{tabular}
\end{table}

From Tab.~\ref{tab:AdjustedPI} the MSIS and mean wQL show that the adjusted PI is sharper than the original PI. The PCIP for the upper and lower boundaries are the same for both PIs and are not reported. 

\section{Conclusions}


This work tackled the problem of making accurate forecasts of RES technologies that has a highly intermittent nature in electricity generation. The increased share of such technologies in the electricity market increases the uncertainties about expected future power generation. To account for such uncertainties, deep learning models can be applied to make probabilistic forecasts. In this work, a day-ahead probabilistic forecast was performed using three different dataset configurations. The deep learning model selected was the DeepAR model which has shown promise in several tasks related to probabilistic forecasts in energy applications. In this work, the DeepAR model was used with different internal units. The prediction performances obtained for each model on the different configurations were compared in terms of obtained quality on the PI. 

The DeepAR model with LSTM units obtained the most accurate prediction performance for all experiments. Among the different configurations, the best performance in terms of sharpness and calibration was obtained when both historical data on measured weather (wind direction + wind speed) \emph{and} the NWPs (wind direction + wind speed) were used as exogenous variables. When using only the predicted weather as exogenous variables, worse results were obtained. This could be due to the highly complex topography of the region where the wind farm is located, which makes it increasingly difficult to make accurate weather forecasts. Using the historical measured wind allows the deep learning model to auto-correct the potential failure sources in the NWPs.

The proposed methodology shows the importance of using measurements data to account for systematic biases in the weather forecasts. As RES technologies are directly affected by the current weather, minimizing the failure sources from weather predictions is something that could provide more accurate forecasts of expected power generation. In the end, using the information about technical limitations of the wind power plant to adjust the upper boundary resulted in a much sharper PI. This illustrates the importance of using physical knowledge about the power system in addition to machine learning techniques. 

\section*{Acknowledgments}
O.F.E, M.C, and F.M.B acknowledge the support from the research project “Transformation to a Renewable \& Smart Rural Power System Community (RENEW)”, connected to the Arctic Centre for Sustainable Energy (ARC) at UiT-the Arctic University of Norway through Grant No. 310026.





\bibliographystyle{abbrv}  
\bibliography{references}  

\end{document}